\documentclass[10pt,journal,compsoc]{IEEEtran}

\pdfoutput=1
\usepackage{times}
\usepackage{latexsym}
\usepackage{graphicx}
\usepackage{booktabs}
\usepackage{multirow}
\usepackage{amsmath}
\usepackage{hyperref}
\usepackage{cleveref}
\usepackage{color}
\usepackage{float}
\usepackage{ulem}
\usepackage{makecell}

\usepackage{microtype}
\usepackage{tabularx}
\newcolumntype{P}[1]{>{\centering\arraybackslash}p{#1}}
\newcolumntype{M}[1]{>{\centering\arraybackslash}m{#1}}

\ifCLASSOPTIONcompsoc
  \usepackage[nocompress]{cite}
\else
  \usepackage{cite}
\fi
\ifCLASSINFOpdf
\else
\fi
\begin{document}
%
\title{Graph-free Multi-hop Reading Comprehension: A Select-to-Guide Strategy}

\author{Bohong~Wu, Zhuosheng Zhang, Hai~Zhao
\IEEEcompsocitemizethanks{
\IEEEcompsocthanksitem This paper was partially supported by National Key Research and Development Program of China (No. 2017YFB0304100), Key Projects of National Natural Science Foundation of China (U1836222 and 61733011), Huawei-SJTU long term AI project, Cutting-edge Machine Reading Comprehension and Language Model. This work was supported by Huawei Noah’s Ark Lab  (Corresponding author: Hai Zhao).
\IEEEcompsocthanksitem B. Wu, Z. Zhang and H. Zhao are with the Department of Computer Science and Engineering, Shanghai Jiao Tong University, and also with Key Laboratory of Shanghai Education Commission for Intelligent Interaction and Cognitive Engineering, Shanghai Jiao Tong University, and also with MoE Key Lab of Artificial Intelligence, AI Institute, Shanghai Jiao Tong University. \protect\\ E-mail: \{chengzhipanpan, zhangzs\}@sjtu.edu.cn, zhaohai@cs.sjtu.edu.cn
}
}

\IEEEtitleabstractindextext{%
\begin{abstract}
Multi-hop reading comprehension (MHRC) requires not only to predict the correct answer span in the given passage, but also to provide a chain of supporting evidences for reasoning interpretability. It is natural to model such a process into graph structure by understanding multi-hop reasoning as jumping over entity nodes, which has made graph modelling dominant on this task. Recently, there have been dissenting voices about whether graph modelling is indispensable due to the inconvenience of the graph building, however existing state-of-the-art graph-free attempts suffer from huge performance gap compared to graph-based ones. This work presents a novel graph-free alternative which firstly outperform all graph models on MHRC. In detail, we exploit a select-to-guide (S2G) strategy to accurately retrieve evidence paragraphs in a coarse-to-fine manner, incorporated with two novel attention mechanisms, which surprisingly shows conforming to the nature of multi-hop reasoning. Our graph-free model achieves significant and consistent performance gain over strong baselines and the current new state-of-the-art on the MHRC benchmark, HotpotQA, among all the published works.
\end{abstract}

}

\maketitle

\IEEEdisplaynontitleabstractindextext

%
\IEEEpeerreviewmaketitle

\ifCLASSOPTIONcompsoc
\IEEEraisesectionheading{\section{Introduction}\label{sec:introduction}}
\else
\section{Introduction}
\label{sec:introduction}
\fi

\begin{figure}
    \centering
    \includegraphics[width=1.0\linewidth]{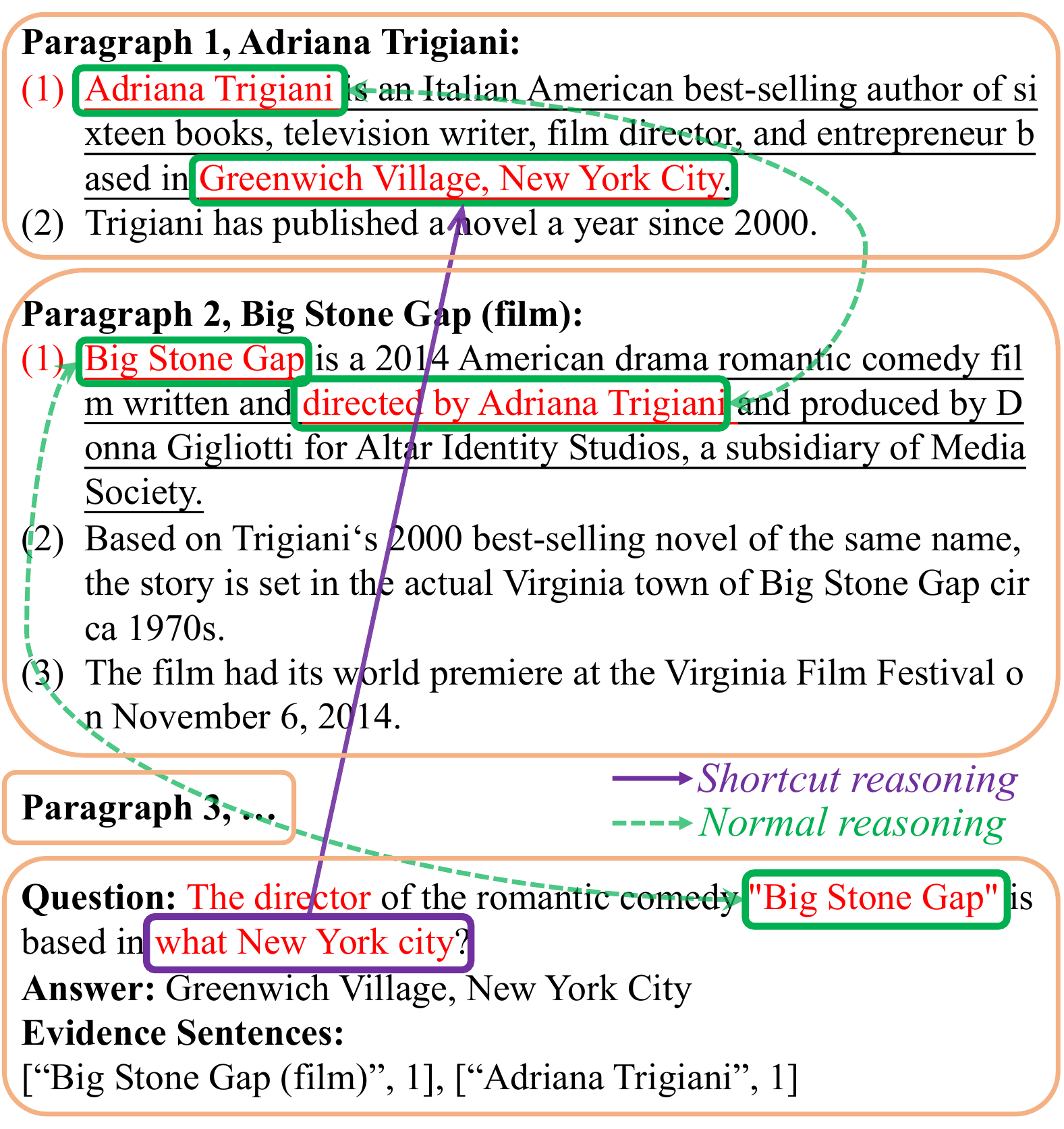}
    \caption{An example of MHRC problem from HotpotQA. In this case, the answer span \textit{Greenwich Village, New York City} could be found in a shortcut as it is closely related to the span \textit{New York City}. Also, finding evidence paragraph \textit{Adriana Trigiani} requires the information from paragraph \textit{Big Stone Gap (film)}.}
    \label{fig:showcase}
\end{figure}

\IEEEPARstart{N}{eural} methods have achieved impressive results on single-hop machine reading comprehension (MRC) datasets \cite{hermann2015teaching, rajpurkar2016squad, trischler2017newsqa, lai2017race, khashabi2018looking}, where questions are usually straightforward and answers are easy to find. As an upgrade, more complicated MRC, multi-hop reading comprehension (MHRC) has to consider reasoning chain in addition to provide an answer \cite{yang2018hotpotqa, welbl2018constructing}, where HotpotQA \cite{yang2018hotpotqa} is one of the most representative MHRC datasets. In detail, MHRC, like other common MRC, also asks model to give an answer according to a given passage (known as \textit{Ans} task in HotpotQA). Meanwhile it is equally important for MHRC tasks to present a series of evidence sentences (known as \textit{Sup} task in HotpotQA) as a proof of multi-hop reasoning at the same time, as there often exist reasoning shortcuts in MHRC \cite{jiang2019avoiding}, shown in Figure \ref{fig:showcase}. Evidence sentences extraction could be seen as a kind of reasoning interpretability for MHRC tasks.

To solve the MHRC task, previous works  \cite{qiu2019dynamically,fang2020hierarchical,ding2019cognitive,tu2019multi, song2018exploring, de2019question, chen2019multihop} have made great progress by incorporating named entity (NE) knowledge and graph modeling, as intuitively, the process of multi-hop reasoning could be interpreted as jumping over entity nodes in the graph. This has made graph-based methods dominant on the HotpotQA benchmark. However, graph modeling usually requires both an NE recognition model to specify NEs in the contexts and meticulous man-made rules to create such a graph. The current best graph-based method Hierarchical Graph Network (HGN) \cite{fang2020hierarchical} even leverages link information from wikipedia pages at scale, which makes graph modelling extremely inflexible for general use.

Researchers recently \cite{groeneveld2020simple, shao2020graph} are questioning about how much this graph structure benefits MHRC. Although the graph-free attempt on MHRC is promising, these graph-free models only obtained sub-optimal performance on the MHRC benchmark, HotpotQA. C2FReader \cite{shao2020graph}, as the previous best graph-free baseline, suffers from huge performance gap compared to the best graph-based model HGN. The performance gap is too significant on the \textit{Sup} task to support any refutation pro graph-free modeling. In consideration of the arguable role for the graph network in MHRC, in this work, we explore effective graph-free modeling to challenge state-of-the-art graph-based methods for MHRC.

In MHRC, one question is often provided with several lexically related paragraphs. It is often infeasible for current pre-trained language models  (PrLMs) to accept all paragraphs as input at one time. Therefore, a paragraph retrieval module \cite{tu2020select, qiu2019dynamically, shao2020graph, fang2020hierarchical} is indispensable to capture the most salient part in the input sequence. However, such module incoordinately attracts little attention in previous studies by simply treating paragraph retrieval as individual sequence classification. This leads to miscapturing two key nature in MHRC. (i) Ranking issue. Evidence paragraph retrieval is essentially a ranking problem, while simple point-wise label (relevant or not in MHRC) is unable to cover the ranking nature \cite{liu2009learning}. (ii) Multi-hop dependency issue. The retrieval of the second evidence paragraph usually depends on the information in previous evidence paragraph (also depicted in Figure \ref{fig:showcase}), while such processing ignores this dependency and only retrieve paragraph within the scope of single-hop. In this paper, to alleviate both of the aforementioned difficulties, we propose our select-to-guide (S2G) model, which reformulates the existing “sequence classification objective” into a “ranking score objective” to capture the ranking nature, and introduces a “score refinement module” to capture multi-hop dependency between paragraphs. Our S2G is further incorporated with two novel attention mechanisms to enhance performance on both subtasks of MHRC. For \textit{Sup} task, we inject a special token before each sentence starts to indicate its individuality, and then introduce \textbf{S}entence-\textbf{a}ware \textbf{S}elf-\textbf{A}ttention (SaSA) to explicitly aggregates all the token embeddings within the sentence. For \textit{Ans} task, motivated by human reading experience which usually pays more attention to evidence sentences \cite{sun2019improving, zhang2020dcmn+}, we introduce \textbf{E}vidence \textbf{G}uided \textbf{A}ttention (EGA) to force model concentrate more on the extracted evidences from previous steps.

In summary, this work presents a “select-to-guide” design which firstly outperforms all graph-based methods by incorporating all the above proposed mechanisms in MHRC modeling (We will release our source code and pretrained models soon). Our contribution in this paper is three-fold:

\noindent$\bullet$ We propose a graph-free S2G model concentrating on a coarse-to-fine, step-wise paragraph selecting method, which accurately pre-extracts the relevant paragraphs to produce a nearly noise-free context for the later comprehension processing.

\noindent$\bullet$ We propose two novel attention mechanisms, which enhance the MHRC model in terms of both $Sup$ and $Ans$ tasks effectively.

\noindent$\bullet$ Our proposed graph-free model S2G achieves the state-of-the-art performance on the MHRC benchmark, HotpotQA, among all the published works, and is the first work that experimentally verifies carefully designed graph-free methods could significantly outperform graph-based ones on MHRC tasks.

\section{Background}


\subsection{Multi-hop Reading Comprehension}


Multi-hop reading comprehension (MHRC), as a special task of Machine Reading Comprehension (MRC), has been a challenging topic in natural language understanding in recent years, and is more closer to real scenarios. To promote the researches on MHRC, there has been many datasets like HotpotQA \cite{yang2018hotpotqa}, WikiHop \cite{welbl2018constructing}, OpenBookQA \cite{mihaylov2018can}, NarrativeQA \cite{kovcisky2018narrativeqa}. HotpotQA is one of the most representative dataset among them, as it not only requires the model to extract the correct answer span from the context, but also to provide a collection of evidence sentences as a proof of MHRC.

For Multi-hop Reading Comprehension, graph based methods were dominant, as it is quite natural to interpretate multi-hop reasoning steps as jumping over entity nodes. DFGN \cite{qiu2019dynamically} simply uses the Stanford corenlp toolkit \cite{manning2014stanford} to extract named entities from the context and query as graph nodes, and then build graph edges between two entity mention if they co-exists in one single sentence. SAE \cite{tu2020select} instead considers sentence pieces as graph nodes, and creates sentence edges based on co-mention of name entities. HGN \cite{fang2020hierarchical} combines the both the strengths of sentence node modeling and entity node modelling, and further incorporates even paragraph nodes in their graph design, introducing a heterogeneous graph neural network, which achieves impressive performance on HotpotQA.

Recently however, there are also some graph-free models like QUARK \cite{groeneveld2020simple} and C2FReader \cite{shao2020graph} questioning about the efficacy of graph modelling. QUARK provides a extremely simple pipeline model that analyzes questions in HotpotQA sentence by sentence, and has achieve acceptable performance close to graph-based methods, taking its simplicity of architecture into consideration. C2FReader, as the previous best graph-free model, theoretically analyzes the difference between graph attention and self attention, and observe negligible performance difference, which indicates that graph modeling is not indispensable for MHRC. However, all these graph-free approaches still suffer from huge performance gap compared to the best graph-based model.

\subsection{Paragraph Retrieval}

For cases in MHRC datasets, one question is usually provided with several lexically related paragraphs,  which contains many redundant while confusing contexts. This makes it infeasible to feed such long inputs into normal pretrained language models at only one time, as BERT-like PrLMs \cite{devlin2018bert} only accept inputs of length 512 tokens. A unique end-to-end solution to this problem is to pretrain totally new language models that accepts long input, like Longformer \cite{beltagy2020longformer} and ETC \cite{ainslie2020etc}. These works are intended to accelerate the process of PrLMs on handling extremely long input, but on the price of performance sacrification and are resurce consuming. Naturally,  cascaded models \cite{qiu2019dynamically, tu2020select, fang2020hierarchical, shao2020graph, groeneveld2020simple} that composed of a reader and a retriever are more often used. These cascaded systems retrieve the most relevant evidence paragraphs first, and perform multi-hop reasoning on retrieved contexts thereafter. 

The mainstream research line on MHRC tasks is to design sophisticated reader module for yielding good performance. In this work, however, we argue that more concentration should be put on the retriever setting. In fact, the retriever is of vital importance in MHRC problems, as fine-grained paragraph retrieval will provide a quasi noise-free context for the following reading comprehension tasks. DFGN\cite{qiu2019dynamically} uses simple BERT-score as the relevance score for each paragraph, which ignores the semantic relation between the paragraphs, thus naturally introduces noise in the selected paragraphs, resulting in poor performance on the reading comprehension tasks. To encourage the interaction between paragraphs and explicitly model the ranking nature of the retrieved paragraphs, \cite{tu2020select} introduces bi-attention between the BERT-scores of each paragraph, and uses pairwise ranking loss for better retrieval performance. Although this method partially covers the ranking nature, it fails to capture the multi-hop dependency between paragraphs, as the retrieval of second-hop paragraph might highly depends on the information in the first hop. HGN \cite{fang2020hierarchical} therefore designs a cascaded paragraph retrieval module, which uses lexical matching for the first hop, and BERT score ranking as the second. However, lexical matching introduces too much noise, making its retrieval performance even worse than SAE.

There also exist other approaches that decompose the multi-hop question into single-hop ones, but such methods either totally ignore the reasoning interpretability task \textit{Sup} \cite{min2019multi, perez2020unsupervised}, or performs sub-optimal in both tasks \cite{nishida2019answering}, and is not the main stream method on MHRC datasets.

\subsection{Sentence Embedding}
To increase the interpretability of the multi-hop reasoning, the HotpotQA dataset also introduces \textit{Sup} task to ask for a complete set of evidence sentences. For \textit{Sup} task, fine-grained sentence embedding is required. Traditional ways of acquiring sentence embedding in MHRC include (i) mean pooling or max pooling over the whole sentence token embeddings and (ii) concatenating the sentence boundary token embeddings. The mean or max pooling strategy are most widely studied and adopted in other research areas that require sentence embeddings\cite{reimers2019sentence, williams2018broad, liu2021filling, tu2020select}. As for the concatenating strategy, some recent works have also pointed out that the sentence boundary tokens are more likely to contain the sentence information \cite{clark2019does}, indicated by the visualization of BERT attention matrix. Therefore, many researches also use this concatenating strategy as an alternative of acquiring sentence embeddings \cite{fang2020hierarchical}.

Recently, there are also some works utilizing a special placeholder token as indicators of a sentence to get sentence embedding. SLM \cite{lee2020slm} injects a “$<$sent$>$” token before each sentence, and introduces “sentence unshuflling” task to pretrain sentence embedding. Longformer \cite{beltagy2020longformer} also injects a $<$s$>$ token, while ETC \cite{zaheer2020big} further creates a sentence token sequence, and design special attention mechanism between the sentence sequence and token sequence to finetune the sentence embeddings. In this work, we follow this injection design to acquire sentence embeddings. Based on this specially injected placeholder token, we further novelly propose sentence-aware self-attention (SaSA) to explicitly constrain these special tokens only attend to the tokens that are within the same sentence, which further enhances the sentence embeddings.

\section{Our Method}
\subsection{Notations}
\begin{figure}[tp]
    \centering
    \includegraphics[width=1.0\linewidth]{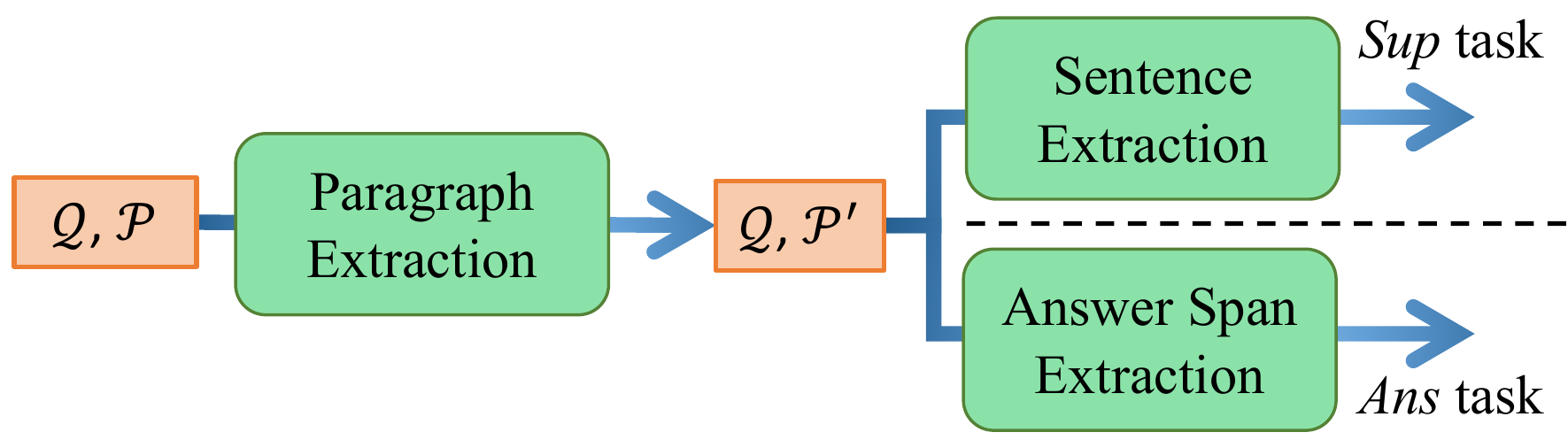}
    \caption{Overview of our processing pipeline on MHRC tasks.}
    \label{fig:overall}
\end{figure}

Formally in MHRC task, each question $\mathcal{Q}$ is usually provided with a paragraph set $\mathcal{P}$, which consists of 10 paragraphs at most, though only a few of them (two in HotpotQA) are truly related to $\mathcal{Q}$. In this paper we take HotpotQA for an example, and deal with it in a pipeline way. An overview of our processing pipeline is shown in Figure. \ref{fig:overall}.

\noindent \textbf{Evidence Paragraph Retrieval}
Given a question $\mathcal{Q}$ and a paragraph set $\mathcal{P}$, our task is to find two most relevant evidence paragraphs' set $\mathcal{P}' = \{\mathcal{P}_1, \mathcal{P}_2\}$. 

\noindent \textbf{Evidence Sentence and Answer Span Extraction} 
Given a question $\mathcal{Q}$ and two corresponding evidence paragraphs, each consists of several sentences $\mathcal{P}_1 = \{s_{1,1}, s_{1, 2}, ..., s_{1, |\mathcal{P}_1|}\}, \mathcal{P}_2 = \{s_{2, 1}, s_{2, 2}, ..., s_{2, |\mathcal{P}_2|}\}$. Our task is to select out all the evidence sentences set $\mathcal{S}^*$ that are related to the question and find the correct answer span within them at the same time. For simplicity, we denote $\mathcal{S}$ as a combination of the sentences from both paragraphs $\mathcal{S} = \{s_{1}, s_{2}, ..., s_{k}\}$, where $k = |\mathcal{P}_1| + |\mathcal{P}_2|$ is the total number of sentences in the paragraphs.

\subsection{Evidence Paragraph Retrieval}

Similar to the Select-Answer-and-Explain (SAE)'s paragraph retrieval strategy \cite{tu2020select}, we first concatenate the question tokens with the paragraph tokens separately for all paragraphs, and use a multi-head self-attention layer to capture the interaction information among these paragraphs. 

\noindent \textbf{Objective Reformulation} Naive paragraph retrieval method uses a 2-way classifier for each paragraph to decide whether it is relevant or not. As we have mentioned above, such design could cover neither the ranking nor the multi-hop nature of MHRC tasks. SAE solves the first issue by adopting a pairwise comparison module for ranking, while our approach handles it by reformulating the sequence classification objective into a score matching objective, which is simpler in use.

With the same assumption as SAE that the paragraph with the exact answer span is more relevant to the question, we heuristically assign a probability score to each paragraph in the paragraph set $\mathcal{P}$, based on whether it is relevant or even contains the exact answer span. Following SAE, we set the score for paragraph that contains the answer to 2, and the second relevant paragraph to 1. We append a softmax operation on the assigned scores to acquire the probability for each paragraph.
\begin{equation} \label{lr1}
score_{p_i}=
\begin{cases}
2& \text{ paragraph $p_i$ has answer} \\
1& \text{ paragraph $p_i$ relevant} \\
0& \text{ paragraph $p_i$ irrelevant}
\end{cases}
\end{equation}
\begin{equation} \label{lr2}
    \hat{P}_{i} = \frac{e^{score_{p_i}}}{\sum_{p_i \in \mathcal{P}} e^{score_{p_i}}}
\end{equation}
where $\mathcal{P}$ is the set of all paragraphs. We then use KL divergence as our training objective for our paragraph retrieval module: 
\begin{equation} \label{kl_obj}
    \mathcal{L}_{para} (P, \hat{P}) = \sum\limits_{i=1}^{10} P_i \log \frac{P_i}{\hat{P}_i} 
\end{equation}
where $P_i$ is the module score output for the $i^{th}$ paragraph. In this way, our method naturally covers the ranking nature without the sophisticated design of pairwise ranking module used in SAE.


\begin{figure*}[tp]
	\centering
	\includegraphics[width=1.0\linewidth]{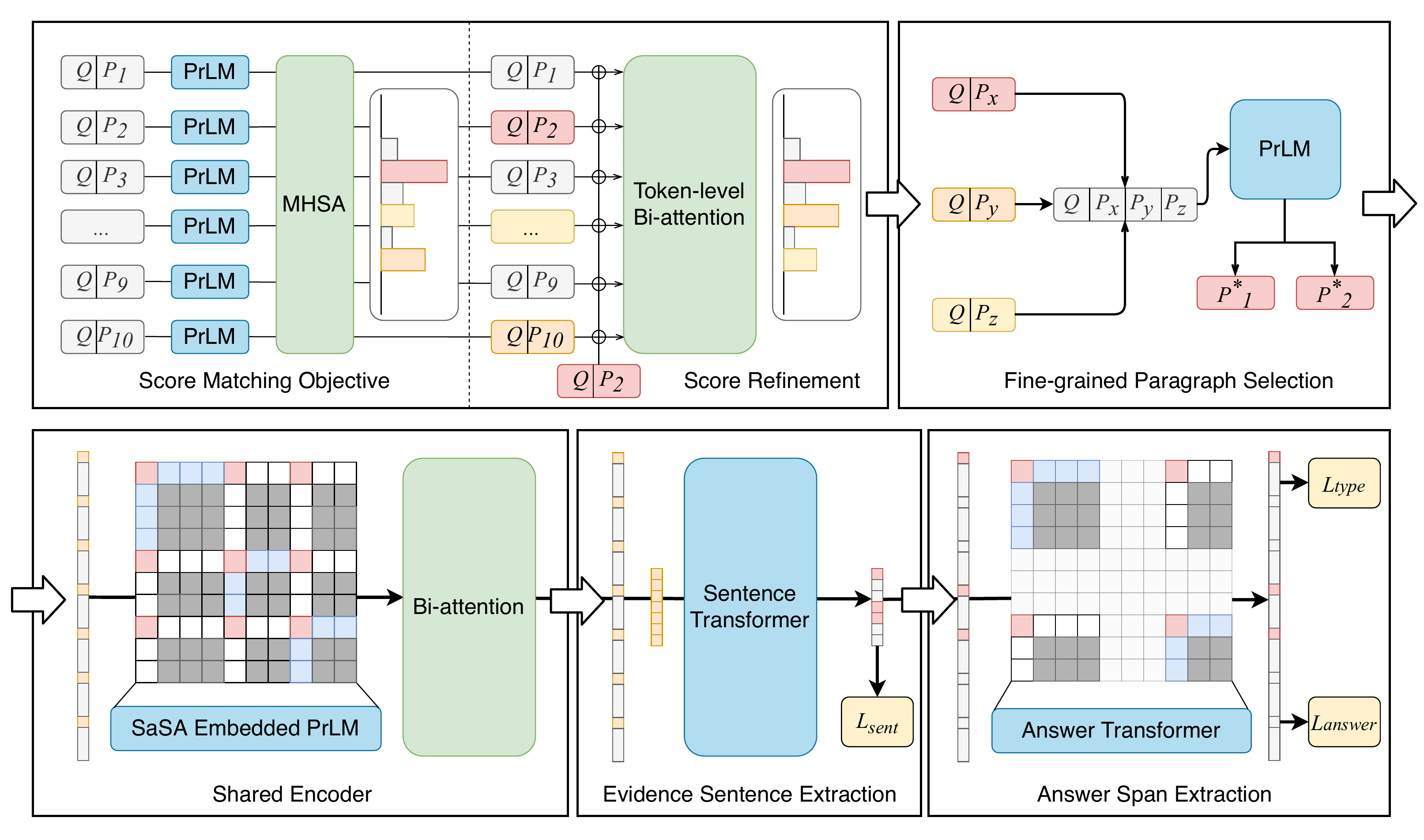}
	\caption{Overview of our S2G model, where (a) the upper flow presents our paragraph retrieval module, and (b) the bottom flow presents our multi-task module. Our paragraph retrieval module consists of three sub-modules, including (i) score matching module, (ii) score refinement module, (iii) fine-grained paragraph retrieval module. Our multi-task module consists of one shared encoder module and two inter-dependent modules targeted on the different subtask, incorporated with our proposed attention mechanisms.}
	\label{fig:pipeline}
\end{figure*}

\noindent \textbf{Score Refinement} The extraction of the second evidence paragraph usually highly depends on the information from the first evidence paragraph, while the extraction in SAE is simply performed in one single hop. Therefore, for the multi-hop retrieval issue, we introduce a “score refinement module” as shown in Figure \ref{fig:pipeline}. At the beginning, we select out the first-hop paragraph $\mathcal{P}_1$, i.e. the paragraph with the highest ranking score after the multi-head self-attention (MHSA) layer. We then take $\mathcal{P}_1$'s sequence output as a new “question vector”, and append a bi-attention layer \cite{seo2016bidirectional} between this “question vector” and all paragraphs, to capture interactions between hops. Two self-attention layers are then appended to aggregate the bi-attention information from the previous stage, followed by a final prediction layer which refines our previous predicted scores for all paragraphs. Note that in HotpotQA, the label for the  first-hop paragraph is not provided, and we simplify this problem by labeling the paragraph with gold answers as the first-hop paragraph.

\noindent \textbf{Cascaded Paragraph Retrieval} 
Considering that the previous retrieval approach neglects the cross attention between the candidate paragraphs, we further adopt another cascaded paragraph retrieval module to select out the relevant paragraphs in a more accurate way. Namely, we choose top $k$ ranking paragraphs in previous retrieval steps and concatenate these paragraphs into a single sequence. We add a “$<$p$>$” token before each paragraph, and feed this sequence into a single BERT model. We use an MLP layer on top of these specially injected indicator tokens to get fine-grained paragraph embeddings for this paragraph subset. Empirically, we set $k$ to $3$ as more paragraphs will likely exceed the length limit by most current PrLMs.

\subsection{Evidence Sentence Extraction}

After acquiring a nearly noise-free context from the paragraph retrieval modules, we concatenate these two paragraphs together to form a new input sequence $\mathcal{S} = \{s_{1}, s_{2}, ..., s_{k}\}$. The following processing on evidence sentence extraction will be based on $\mathcal{S}$. Similar to Longformer \cite{beltagy2020longformer}, we inject a special placeholder token “$<$e$>$” before each sentence to generate sentence embeddings. Then a PrLM such as BERT \cite{devlin2018bert}, RoBERTa \cite{liu2019roberta}, ELECTRA \cite{clark2020ELECTRA} or ALBERT \cite{lan2019albert} will be adopted to obtain contextualized representation for the concatenated input sequence $(\mathcal{Q}, \mathcal{S})$.

\begin{figure}[tp]
    \centering
    \includegraphics[width=1.0\linewidth]{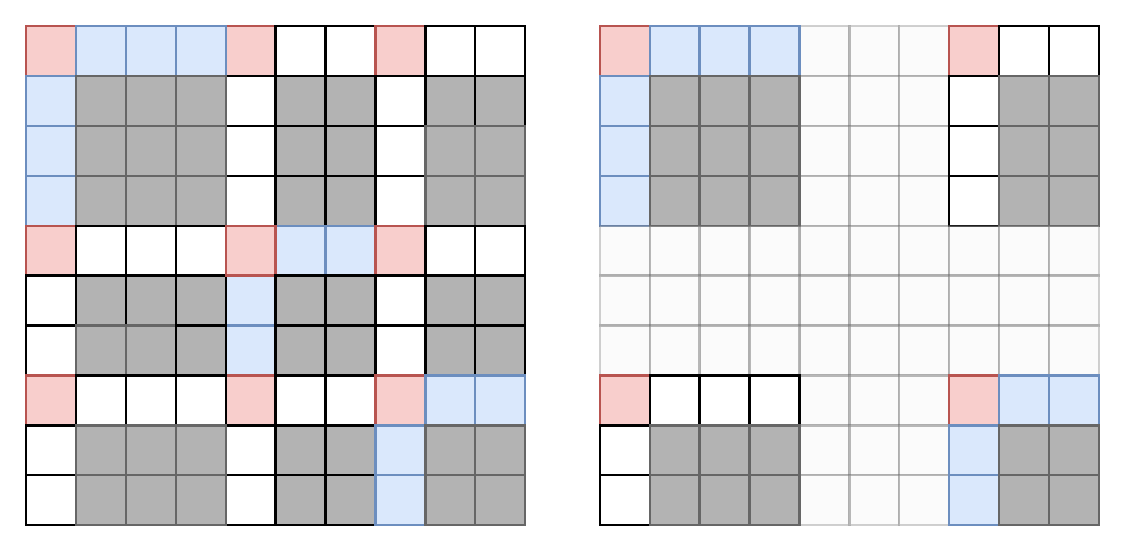}
    \caption{Minimum examples of our proposed SaSA mechanism (left) and our EGA mechanism (right). The original tokens remain fully attended (gray grids). For new injected “$<$e$>$” tokens, each only attends to the corresponding sentence tokens (blue grids) and other “$<$e$>$” tokens including itself (red grids). For EGA, tokens not in evidence sentences are directly masked out (translucent grids).}
    \label{fig:attention}
\end{figure}

\noindent \textbf{Sentence-aware Self-Attention} Here we denote all “$<$e$>$” tokens as $\mathcal{T}_s$, and other tokens as $\mathcal{T}_w$. It is obvious that $\mathcal{T}_s \cap \mathcal{T}_w = \emptyset$ and $\mathcal{T}_s \cup \mathcal{T}_w = \mathcal{T}$. Define $g (t_i, t_j)$ as a binary function on $t_i$ and $t_j$, and $g (t_i, t_j) = 1$ stands only when $t_i, t_j$  belongs to the same sentence $s^\prime$. With the straight-forward assumptions that the sentence placeholder tokens should only attend and be attended to the tokens that are within their corresponding sentences, we integrate this attention scheme into PrLMs by designing the following masking strategy:
\begin{equation} \label{sasa}
    M_{SaSA}[i, j]=\left\{
\begin{array}{cc}
0      &   {if\ t_i, t_j \in \mathcal{T}_s}\\
0    &   {if\ t_i, t_j \in \mathcal{T}_w}\\
0     &   {if\ g (t_i, t_j) = 1}\\
-\infty  &   {otherwise}
\end{array} \right.
\end{equation}
A more specific view of our attention mechanism is shown in Figure \ref{fig:attention}. With such design of the attention mechanism, each sentence placeholder token has softly gathered the embedding information from all tokens in its sentence, thus is a more natural way to acquire sentence embeddings. For simplicity, we only consider at most the first $k$ sentences in the contexts, where $k$ should be a hyperparameter that covers most of the cases in HotpotQA. In our experiments we assign $k$ to $14$, as it already covers over $99\%$ of the cases.


\noindent \textbf{Sentence Transformers} We first extract out all the sentence placeholder tokens to form a new sequence of length $k$. Considering that question sentence embedding is useful for reasoning, we append the question placeholder token to the end of this sequence. To enrich the bi-directional interaction between the extracted placeholder tokens, we further append $t$ self-attention layers on the new sequence, denoted as our “Sentence Transformer” module. A one-layer MLP is followed to get the final sentence prediction results $o_{sent}$.

\subsection{Answer Span Extraction}  \label{span_extract}
With predictions of supporting evidences provided in the previous evidence extraction module, we introduce evidence guided attention (EGA) to help the answer span extraction. Here we constrain the tokens that are relevant to the questions to only attend and be attended to themselves, making them “unseen” from the irrelevant contexts (see Figure \ref{fig:attention} for details). Let the sentence prediction result from previous module be $Z = \{z_1, z_2, z_3, ..., z_k\}$, where $z_i = 1$ means sentence $s_i$ is selected as the supporting evidence and $z_i=0$ means the opposite. Let $\sigma (\cdot)$ be a mapping from tokens to sentences, where $\sigma (t_i) = j$ means that token $t_i$ belongs to sentence $s_j$. Our evidence guided attention (EGA) mechanism is then implemented by designing the following masking strategy:
\begin{equation} \label{token_eq}
    M_{EGA}[i, j]=\left\{
\begin{array}{ccc}
-\infty      &      & {if\ z_{\sigma (t_i)} = 0}\\
-\infty      &      & {if\ z_{\sigma (t_j)} = 0}\\
0    &      & {otherwise}\\
\end{array} \right. 
\end{equation}

\noindent \textbf{Answer Transformers} Similar to our “Sentence Transformer” component, we also append $t$ self-attention layers embedded with EGA mechanism to perform more fine-grained answer span extraction, denoted as our “Answer Transformer” component. A two-layer MLP is appended upon this component to get the final predictions of the start logits $o_{start}$ and end logits $o_{end}$ of the answer spans. As there also exist a number of questions whose answers are either “yes” or “no”,  we use a two-layer MLP on the representation of the first “$<$s$>$” token to get type prediction result $o_{type}$. 

\subsection{Multi-task Prediction} \label{predict}
In consistent with \cite{qiu2019dynamically, tu2020select, fang2020hierarchical}, we treat both \textit{Ans} and \textit{Sup} tasks into a multi-tasking framework. Our total loss function is a combination of three CrossEntropy loss based on all previous logits $o_{sent}$, $o_{start}$ and $o_{end}$:
\begin{equation} \label{loss_term}
    \mathcal{L}_{joint} = \lambda_1  \mathcal{L}_{sent} + \lambda_2  \mathcal{L}_{span} + \lambda_3 \mathcal{L}_{type} 
\end{equation}
where $\lambda_1$, $\lambda_2$ and $\lambda_3$ are adjustable hyperparameters. Empirically, we set $\lambda_1$ to $2$ and other parameters to $1$.

\begin{table*}[tp]
	\caption{Blind test results on the distract setting of HotpotQA. Bold statistics means the best metrics over all models, and underlined statistics represents the best metrics among published graph-based methods. Our results achieve overall the best performance on HotpotQA's benchmark, and outperforms previous best published graph-based model HGN by a significant margin.} 
	\label{tab:results}
	\setlength{\tabcolsep}{20.0pt}
	\begin{tabular}{ll|cccccc}
		\toprule  
		\multirow{2}{*}{Model}& & \multicolumn{2}{c}{Ans}&\multicolumn{2}{c}{Sup} & \multicolumn{2}{c}{Joint}\\
		& & EM & F1 & EM & F1 & EM & F1 \\
        \midrule
        \multicolumn{8}{l}{\textit{Other methods}} \\
		DecompRC & \cite{min2019multi} & 55.20 & 69.63 & - & - & - & - \\
		ChainEx & \cite{chen2019multihop} & 61.20 & 74.11 & - & - & - & - \\
		QFE & \cite{nishida2019answering} & 53.86 & 68.06 & 57.75 & 84.49 & 34.63 & 59.61 \\
		
		\midrule
		\multicolumn{8}{l}{\textit{Graph-based methods}} \\
		
		DFGN & \cite{qiu2019dynamically} & 56.31 &	69.69	& 51.50	& 81.62	& 33.62 &	59.82 \\
		
		SAE & \cite{tu2020select} & 66.92 & 79.62 & 61.53 & 86.86 & 45.36 & 71.45 \\
		
		HGN & \cite{fang2020hierarchical} & \uline{69.22} & \uline{82.19} & \uline{62.76} & \uline{88.47} & \uline{47.11} & \uline{74.21} \\
		
		\midrule

		\multicolumn{8}{l}{\textit{Graph-free methods}}  \\
		C2FReader & \cite{shao2020graph}  & 67.98 & 81.24 & 60.81 & 87.63 & 44.67 & 72.73 \\
		
		Longformer & \cite{beltagy2020longformer} & 68.00 & 81.25 & 63.09 & 88.34 & 45.91 & 73.16 \\
		
		ETC & \cite{ainslie2020etc} & 68.12 & 81.18 & 63.25 & \textbf{89.09} & 46.40 & 73.62 \\
		
		
		S2G & Ours & \textbf{70.72} & \textbf{83.53} & \textbf{64.30} & 88.72 & \textbf{48.60}  & \textbf{75.45} \\
		
		\multicolumn{2}{l}{S2G vs. HGN} & $\uparrow 1.50$ & $\uparrow 1.34$ & $\uparrow 1.64$ & $\uparrow 0.25$ & $\uparrow 1.49$ & $\uparrow 1.24$ \\

		\bottomrule
	\end{tabular}
\end{table*}

\section{Experiments}
In this section, we will compare our model with SAE and other strong baselines on the HotpotQA benchmark in detailed analysis. Unless specially noted, we use RoBERTa \cite{liu2019roberta} as our PrLM.

\subsection{Dataset}
HotpotQA  \cite{yang2018hotpotqa} is the first dataset that introduces multi-hop machine reading comprehension. In this paper we focus on the distractor setting of HotpotQA, following the same research line as previous studies \cite{qiu2019dynamically, tu2020select}. In this setting, each question is usually provided with 10 paragraphs, where one system has to solve the following two subtasks simultaneously.

\noindent \textbf{Supporting evidence extraction (\textit{Sup})} For each question, one system has to extract all the evidence sentences among all paragraphs as a proof of reasoning interpretability.

\noindent \textbf{Answer span extraction task (\textit{Ans})} For each question, one system has to extract the most likely answer span among all paragraphs. 

Exact Match (EM) score and F1 score are used to evaluate the performance. The HotpotQA benchmark also considers the EM and F1 scores of a joint prediction on both tasks.

\subsection{Implementation Details}

\noindent\textbf{Cascaded Paragraph Retriever Module}
For paragraph retrieval module implementation, we adopt RoBERTa \cite{liu2019roberta} as our baselines for each sub module in this cascaded module. We vary the initial learning rate in \{1e-5, 2e-5, 3e-5\} with a warm-up rate of 0.1 and L2 weight decay of 0.01. The batch size was selected in \{8, 16, 32\}. The maximum number of epochs was set to 8. We use wordpieces to tokenize all the texts, and the maximum input length was set to 512. The configuration for multi-head self-attention was the same as that for BERT.

\noindent\textbf{Reading Comprehension Module}
For reading comprehension module implementation, we adopt BERT \cite{devlin2018bert}, RoBERTa \cite{liu2019roberta}, ELECTRA \cite{clark2020ELECTRA} and ALBERT \cite{lan2019albert} as our baselines. Similarly, we vary the initial learning rate in \{1e-5, 2e-5, 3e-5\} on the base settings of PrLMs, and in \{5e-6, 1e-5, 2e-5\} on the large settings. Due to resource limitation, we only vary the batch size in \{8, 16\}. Each setting on BERT, RoBERTa, ELECTRA was trained on 4 Tesla P40 GPUs for at most 6 epochs. It takes 1.5 hour per epoch on the base settings of BERT, RoBERTa, ELECTRA, and 3 hour per epoch on the large settings of BERT, RoBERTa, ELECTRA. For ALBERT-xxlarge, it takes neraly 5 hours per epoch on 8 Tesla P40 GPUs, and we only train it for at most 4 epochs.

\subsection{Experimental Results}
\noindent \textbf{Paragraph Retrieval} For evidence paragraph retrieval, we compare our method with two state-of-the-art graph based models, HGN \cite{fang2020hierarchical} and SAE \cite{tu2020select}, which are also implemented in a Retriever + Reader pipeline. Results are shown in Table  \ref{tab:para_result}. In table “Gold” means selected paragraphs contain the answer span. Metrics are calculated based on the top-2 ranked paragraphs selected by each model. As for single model retrieval performance, our model has already surpassed previous baselines significantly on all metrics. Even with smaller PrLMs compared to SAE, our base model achieves 1.5\% absolute improvement (93.43\% EM vs. 91.98\% EM). Our paragraph retriever module yields even higher performance when the cascaded setting is incorporated, outperforming previous best model by large margin (+3.79\% EM).

\begin{table}[tp]
	\centering
    \caption{Paragraph retrieval results on the dev set of HotpotQA. “-” means not provided in their paper.}
    \label{tab:para_result}
    \setlength{\tabcolsep}{15.0pt}
    \begin{tabular}{l|ccc}
        \toprule
        \makecell[c]{Model} & EM & F1 & Gold \\
        \midrule
        SAE$_{large}$ (rerun) & 91.98 & 95.76 & 97.03  \\
        HGN & - & 94.53 & - \\
        \midrule
        Ours$_{base}$ single & 93.43 & 96.54 & 97.02 \\
        Ours$_{large}$ single & 93.81 & 96.82 & 97.64\\
        Ours$_{large}$ cascaded & \textbf{95.77} & \textbf{97.82} & \textbf{98.47}\\
        \bottomrule
    \end{tabular}
\end{table}

\noindent \textbf{Reading Comprehension on HotpotQA} We compare our S2G model with strong baselines in Table  \ref{tab:results}\footnote{We list the results of current best-performing models from \href{https://hotpotqa.github.io/}{the HotpotQA leaderboard}.}, which shows that our S2G achieves statistically significant better performance on all metrics consistently, with an average gain of over 1.0\%. Besides, our S2G model not only greatly outperforms previous best graph-free model C2FReader by significant large margin (both Longformer and ETC pretrain new language models thus their performances are not directly comparable, however, our S2G still manages to surpass their performances), but also outperforms any graph-based models, reaching new state-of-the-art in the leaderboard among all published works.

\subsection{Ablation Study} 

\noindent \textbf{Paragraph Retrieval Module} To study the effects of each component in our paragraph retrieval module, we conduct two ablation studies on our score matching objective and score refinement module (the effectiveness of our FPS module is already depicted in Table  \ref{tab:para_result}). We try to select out the corresponding paragraphs only in one single-hop (- refine) or simply reformulates the score matching objective back to sequence classification objective (- reformulation). The results are shown in Table  \ref{tab:paraabl}. We could see from the table that the objective reformulation strategy is of vital importance in our paragraph retrieval module, which accounts for nearly 2.0\% performance gain on the EM metric. The score refinement strategy can further significantly improve the paragraph retrieval result by 0.7\% EM. All these results indicate that our proposed paragraph retrieval module is well capable of exploiting the ranking nature as well as the multi-hop dependency between paragraphs.

\begin{table}[tp]
    \centering
    \caption{Ablation on paragraph retrieval module on dev set of HotpotQA. }
    \label{tab:paraabl}
    \setlength{\tabcolsep}{13.0pt}
    \begin{tabular}{l|ccc}
        \toprule
        \makecell[c]{Model} & EM & F1 & Gold \\
        \midrule
        Full (RoBERTa$_{base}$)& \textbf{93.4} & \textbf{96.5} & \textbf{97.0} \\
        \ \ -\ refine & 92.7 & 96.3 & 96.8\\
        \ \ -\ refine -\ reformulation & 90.7 & 95.2 & 95.8 \\
        \bottomrule
    \end{tabular}
\end{table}

\noindent \textbf{Reading Comprehension Module} We conduct ablations on each component in our multi-tasking evidence sentence and answer span extraction. Results are shown in Table  \ref{tab:decoding_abl}. (i) For $Sup$ task, removing our SaSA component will result in small performance drop on both “Sup EM” ($\downarrow 0.2\%$ EM) and “Joint EM” ($\downarrow 0.2\%$ EM) metric, indicating the necessity of SaSA for \textit{Sup} task. (ii) Also, removing our Sentence Transformer component will result in significant performance drop on all metrics, “Ans EM” ($\downarrow 0.3\%$ EM), “Sup EM” ($\downarrow 0.4\%$ EM) metric and “Joint Em” ($\downarrow 0.5\%$ EM) metric. (iii) For $Ans$ task, removing our EGA-based Answer Transformer component will result in a $0.5\%$ EM drop on “Ans EM” metric. All these results indicate the the indispensability of our proposed methods to the overall MHRC performance.

\noindent \textbf{Single Task Ablation}
To verify the dependency between these two subtasks on HotpotQA, we train our model on either single task. The results are show in Table  \ref{tab:decoding_abl}. Removing out either of the subtask will significantly hurt the other subtask performance, which verifies the complementarity of both subtasks in model design.

\noindent \textbf{Entity Graph Ablation}
We incorporate an entity graph into our model to see whether graph structures could furthermore help boost the performance on our current model. We follow HGN's graph design and re-implement a simplified entity graph model. Unlike HGN, we only consider edges between sentence nodes and entity nodes without any paragraph nodes. This is because we only extract the two most relevant evidence paragraphs while HGN chooses more paragraphs for better recall. Results are also shown in Table  \ref{tab:decoding_abl}. In our experiment, however, we didn't witness any performance gain when graph structures are incorporated, but even a slight degradation on both tasks.

\begin{table}[tp]
    \centering
    \caption{EM ablation on two subtasks of HotpotQA on dev set.}
    \label{tab:decoding_abl}
    \setlength{\tabcolsep}{14.0pt}
    \begin{tabular}{l|ccc}
    \toprule
    \makecell[c]{Model}& Ans & Sup & Joint \\
    \midrule
    Full (RoBERTa$_{base}$)& \textbf{65.1}   & \textbf{63.5}   &  \textbf{44.5}    \\
    \ \ -\ SaSA &  \textbf{65.1}  &  63.3  & 44.3 \\
    \ \ -\ Sentence Transformer  &  64.8  &  63.1  &  44.0 \\
    \ \ -\ Answer Transformer  &  64.6  &  63.4  & 44.0 \\
    \midrule
    \ \ -\ \textit{Sup} task & 64.1 & - & -\\
    \ \ -\ \textit{Ans} task & - & 61.7 & - \\
    \midrule
    \ \ +\ Entity Graph & 64.7 & 63.1 & 44.0 \\
    \bottomrule
    \end{tabular}
\end{table}

\noindent \textbf{Different PrLMs} Table  \ref{tab:diff_pre-train} presents the results of employing different PrLMs as backbone, showing that our S2G model achieves consistent performance gain in all fair comparison environments with all other methods. On “Sup-EM” metric and “Joint-EM” metric, especially, our S2G achieves over 1.5\% EM gain in all comparisons with other methods when the same PrLM is used, which further indicates the effectiveness of our proposed method on reasoning interpretability.

\begin{table}[tp]
    \centering
    \caption{EM ablation on different pre-trained models on the dev set. $\dagger$ means the blind test results, for C2FReader didn't report the dev results in their paper.}
    \label{tab:diff_pre-train}
    \setlength{\tabcolsep}{9pt}
    \begin{tabular}{l|l|ccc}
    \toprule
    \makecell[c]{Model} & \makecell[c]{PrLM} & Ans & Sup & Joint\\
    \midrule
    DFGN & BERT$_{large}$  & 55.7  & 53.1  &  33.7 \\
    C2FReader$\dagger$ & RoBERTa$_{large}$ & 68.0 &  60.8 & 44.7 \\
    SAE & RoBERTa$_{large}$ & 67.7  & 63.3  & 46.8 \\
    HGN & ALBERT$_{xxlarge}$ & 70.2 & 63.2 &  47.0 \\
    \midrule
    \multirow{4}{*}{Our S2G} & BERT$_{large}$ & 64.1 & 62.8 & 44.1 \\
        & RoBERTa$_{large}$ & 68.5 & 64.9 & 47.7 \\
        & ELECTRA$_{large}$ & 70.3 & 65.1 & 48.5 \\
        & ALBERT$_{xxlarge}$ & \textbf{70.8} & \textbf{65.7} & \textbf{49.6}  \\
    \bottomrule
    \end{tabular}
\end{table}

\section{Discussion}
\subsection{The efficacy of our paragraph retrieval method}

To further verify that multi-hop paragraph retrieval is a crucial part for MHRC, we conduct experiments to observe the final performance difference on different paragraph retrieval methods. The results are shown in Table  \ref{tab:diff_para}, where all other modules are kept the same except for the paragraph retrieval module. Here “Oracle” means we directly use the paragraphs annotated by the dataset, which represents the upper bound for any paragraph retrieval module. From the table we see that our S2G paragraph retrieval module achieves impressive performance gain on both subtasks compared with SAE paragraph retrieval module. More importantly, our paragraph retrieval nearly reaches the full potential of both subtasks. For \textit{Ans} task, our S2G achieves 98.3\% oracle performance, and for \textit{Sup} task, our S2G achieves 96.6\% oracle performance, indicating the extreme effectiveness of our S2G paragraph retrieval module.

\begin{table}[tp]
	\centering
	\caption{EM ablation on different paragraph retrieval methods on the dev set of HotpotQA.}
	\label{tab:diff_para}
	\setlength{\tabcolsep}{10.0pt}
	\begin{tabular}{l|ccc}
		\toprule
		\makecell[c]{Extraction Methods} & Ans & Sup & Joint \\
		\midrule
		Oracle &  72.0 &  68.0 &  51.2 \\
		\midrule
		SAE extraction  &  69.4 &  62.9 & 47.4  \\
		Our S2G w/o cascaded retriever & 69.9  & 64.1  & 48.2  \\
		Our S2G w. cascaded retriever  & \textbf{70.8} & \textbf{65.7} & \textbf{49.6} \\
		\bottomrule
	\end{tabular}
\end{table}

\subsection{The effect of sentence embedding}

\begin{table}[tp]
	\centering
	\caption{Different methods of capturing sentence embedding}
	\label{tab:diff_sent_method}
	\setlength{\tabcolsep}{11.5pt}
	\begin{tabular}{l|ccc}
		\toprule
		\makecell[c]{Methods} & Ans & Sup & Joint \\
		\midrule
		Mean pooling & \textbf{65.1} & 63.2 & 44.4\\
		Concatenating start and end & 64.0 & 63.0	& 44.0 \\
		Special token  $<$s$>$	& 64.9	& 63.1	& 44.2 \\
		Special token  $<$e$>$ & \textbf{65.1}	& \textbf{63.5}	& \textbf{44.5} \\
		\bottomrule
	\end{tabular}
\end{table}

For the \textit{Sup task} of HotpotQA, acquiring a fine-grained sentence embedding is inevitable in boosting the performance. In this section, we compare different methods of generating sentence embedding.

\noindent\textbf{Mean pooling} Mean pooling is a widely used strategy in getting the sentence embedding \cite{tu2020select, reimers2019sentence, liu2021filling}. In our experiments, we extracts all the tokens in each sentence and average their embeddings as the sentence embedding.

\noindent\textbf{Start token + End token} Studies have found out that the start and the end token are most likely to contain the contextualized sentence embedding \cite{clark2019does}. Therefore, concatenating the start token and end token embedding as a segment's embedding is also widely used in a wide range of works \cite{fang2020hierarchical, lee2020learning}. In our experiments, we concatenate these token embedding together, which results in doubled dimension space, and then use a dense layer to project the concatenated embedding back to the original dimension space.

\noindent\textbf{Special token $<$s$>$} An alternative way of our S2G model is to use the predefined token $<$s$>$ in original PrLMs to indicate each sentence. In our experiments, we also implement it for performance comparison.


The results are shown in Table \ref{tab:diff_sent_method}. In our experiments, we find that using the special token $<$e$>$ as the indicator of each sentence perform overall the best among all the methods. Compared to another special token design, $<$s$>$, especially, our method achieves slightly better performance. We speculate this is because the prediction of the other sub-task in HotpotQA also depends on the embedding of $<$s$>$ token. Using the same token as indicators for both tasks would result in performance interference. Therefore, a totally new token $<$e$>$ would be a better choice for acquiring the sentence embeddings.

\subsection{Error Analysis} \label{error_anl}

\begin{table*}
	\centering
	\caption{Error Case Study of Our S2G model.}
	\label{tab:case_study}
	\setlength{\tabcolsep}{5.0pt}
	\begin{tabularx}{\textwidth}{c|X}
		\toprule
		Error Type & Example Case \\
		\midrule
		
		Answer Incompleteness &  \parbox[c]{\hsize}{\textbf{Paragraph 1}: Basketball at the 2015 Pacific Games \newline
		\uline{(1) Basketball at the 2015 Pacific Games in \textcolor{red}{Port Moresby, Papua New Guinea} was held at the BSP Arena and PNG Power Dome on 3\u201312 July 2015.}
\newline
		\textbf{Paragraph 2}: Kiribati national basketball team \newline
		(1) The Kiribati national basketball team are the basketball side that represent Kiribati in international competitions.
		\uline{(2) They competed at the 2015 Pacific Games, where they finished with an 0-4 record.}
\newline
		\textbf{Question}: What city and nation was the location of the 2015 Pacific Games where the Kiribati national basketball team finished with a 0-4 record?
\newline
        \textbf{S2G Retrieved Paragraph}: 1. Basketball at the 2015 Pacific Games 2. Kiribati national basketball team \newline
		\textbf{Correct Answer}: Port Moresby, \textcolor{red}{Papua New Guinea}
 \newline
		\textbf{S2G Prediction}: Port Moresby} \\
		
		\midrule
		
		Answer Superfluity & \parbox[c]{\hsize}{\textbf{Paragraph 1}: Peabody Hotel \newline
		(1) The Peabody Memphis is a luxury hotel in Downtown Memphis, Tennessee.
		\uline{(2) The hotel is known for the "Peabody \textcolor{red}{Ducks}" that live on the hotel rooftop and make daily treks to the lobby.
}
		(3) The Peabody Memphis is a member of Historic Hotel of America the official program of the National Trust for Historic Preservation. \newline
		\textbf{Paragraph 2}: Hyatt Regency Orlando \newline
		(1) The Hyatt Regency Orlando is a hotel directly connected to the Orange County Convention Center in Orlando, Florida.
		\uline{(2) The 32-story, 1641-room hotel was originally constructed in 1986 as The Peabody Orlando, a brand extension of the original Peabody Hotel in Memphis, Tennessee.
}\newline
        \textbf{S2G Retrieved Paragraph}: 1. Peabody Hotel 2. Hyatt Regency Orlando \newline
		\textbf{Question}: The Peabody Hotel in Memphis (and a sister hotel in Orlando) are named after what residents of the hotel rooftop?
\newline
		\textbf{Correct Answer}: Ducks \newline
		\textbf{S2G Prediction}:  \textcolor{red}{Peabody} Ducks}
\\
		
		\midrule 
		Multi-hop Reasoning & \parbox[c]{\hsize}{\textbf{Paragraph 1}: Budget Rent a Car \newline
		\uline{(1) \textcolor{red}{Budget Rent a Car System}, Inc. is an American car rental company that was founded in 1958 in Los Angeles, California by Morris Mirkin.}
		(2) Budget's operations are headquartered in Parsippany-Troy Hills, New Jersey.
\newline
		\textbf{Paragraph 2}: ACRISS Car Classification Code \newline
		\uline{(1)The ACRISS Car Classification Code developed and
maintained by ACRISS (the Association of Car Rental Industry Systems Standards) and is designed to enable customers and travel professionals to make an informed choice when booking car rental in Europe, Middle East and Africa.
}
		\uline{(2) ACRISS Members include Avis, \textcolor{red}{Budget}, Alamo, National, Enterprise, Europcar, \textcolor{red}{Hertz} and Maggiore}
\newline
		\textbf{Question}: Which American car rental company is also a member of the Association of Car Rental Industry Sytems Standards?
 \newline
        \textbf{S2G Retrieved Paragraph}: 1. ACRISS Car Classification Code \textcolor{red}{2. The Hertz Corporation} \newline
		\textbf{Correct Answer}: Budget Rent a Car\newline
		\textbf{S2G Prediction}:  Hertz Corporation} \\
		\bottomrule  
	\end{tabularx}
\end{table*}

To have a clear clue of what type of errors our current S2G model will make on the reading comprehension tasks of HotpotQA dataset for future improvement, we conduct a simplified error analysis on our best performing S2G model. We roughly classify the acquired error cases into the following error types shown in Table  \ref{tab:error_percent}. All error cases occupy about 29.2\% (1 - 70.8\%) of the entire dev set on HotpotQA.
For our concerned error type “Multi-hop Reading Comprehension”, it occupies only 25.5\% of the total error cases, which only accounts for 7.4\% of the whole dev set. Most of the errors are caused due to either incompleteness or superfluity of the predicted answer spans. This phenomenon indicates that our S2G model is already well capable of capturing multi-hop dependency in the context.

\begin{table}[tp]
	\centering
	\caption{The ratio of each error cases from our S2G.}
	\label{tab:error_percent}
	\setlength{\tabcolsep}{12.0pt}
	\begin{tabular}{c|ccc}
		\toprule
		Kind &  Incomplete & Superfluous & Multi-hop RC\\
		\midrule
		Ratio & 42.38\% & 32.17\% & 25.45\%\\
		\bottomrule
	\end{tabular}
\end{table}

We further provide concrete examples in Table  \ref{tab:case_study} to provide a more detailed illustration of these possible error categories. For error cases in  “Answer Incompleteness” and “Answer Superfluity”, we can see that our S2G could already locate the answer span, but only fail to predict the correct boundary. This fuzzy boundary problem is a common and unsolved problem \cite{yang2020multi}, which is not the main focus of this paper. For “Multi-hop reasoning”, from Table \ref{tab:case_study}, we can see that the error is made because our S2G fails to retrieve the correct paragraph “Budget Rent a Car”, while instead it retrieves a very confusing paragraph “The Hertz Corporation” that co-exits with the correct answer in the same sentence. This phenomenon indicates that there is still space for the improvement of multi-hop paragraph retrieval, which we leave for future work.
%
%

\section{Conclusion}

This paper proposes a novel “select-to-guide” model (S2G) for multi-hop reading comprehension in more effective and convenient way. As an alternative of the existing graph modeling, the proposed graph-free S2G model consists of an evidence paragraph retrieval module which selects evidence paragraphs in a step-by-step multi-hop manner, and a multi-task module that simultaneously extracts evidence sentences and answer spans.

For the paragraph retrieval module, this paper introduces a cascaded paragraph retrieval module that retrieve the evidence paragraphs in a explicit coarse-to-fine manner. Taking both the multi-hop dependency and the ranking nature between paragraphs into consideration, experiments on the HotpotQA shows that our cascaded module achieves significantly better performance than previous single-stage retrieval methods. For the multi-task reading comprehension module, this paper introduces two kinds of task-specific attention mechanisms to restrict the receptive fields of each tokens according to the natures of each specific tasks. Experiments also prove the efficacy of the proposed attention mechanisms. 

For long time, the research line on multi-hop reading comprehension focus on exploiting the multi-hop nature on the reader module by graph modelling, while ignoring the multi-hop nature may also exists in the paragraph retrieval steps. This work first points out that it is the retrieval stage that researchers should pay more attention to. Unlike previous approaches that focus on designing complicated graph-based reader module structures, our S2G could achieve better performance with much simpler reader module design. Concrete error analysis on our current S2G model shows that there is still room for improvement on the multi-hop retriever module design. 

In summary, this work provides an insight for multi-hop reading comprehension, which novelly models the multi-hop reasoning nature in the retriever module instead of the reader module. We hope this work can help further facilitate related researches in the multi-hop reading comprehension area, and shed lights on the topic whether graph modeling is necessary, especially in the blowout of the graph-based designs.


%





\ifCLASSOPTIONcaptionsoff
  \newpage
\fi



\normalem
\bibliographystyle{IEEEtran}
\bibliography{anthology.bib}

\vspace{-12mm}
	\begin{IEEEbiography}[{\includegraphics[width=1in,height=1.25in,clip,keepaspectratio]{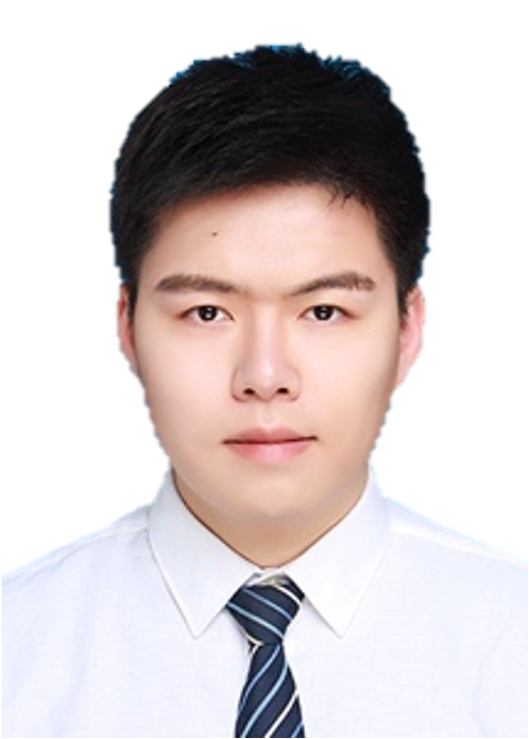}}]{Bohong Wu}
  		received his Bachelor's degree in Computer Science from Shanghai Jiao Tong University in 2018. He is working towards the M.S. degree in computer science with the Center for Brain-like Computing and Machine Intelligence of Shanghai Jiao Tong University. His research interests include natural language processing, machine reading comprehension, and machine translation. 
	\end{IEEEbiography}

\vspace{-12mm}
	\begin{IEEEbiography}[{\includegraphics[width=1in,height=1.25in,clip,keepaspectratio]{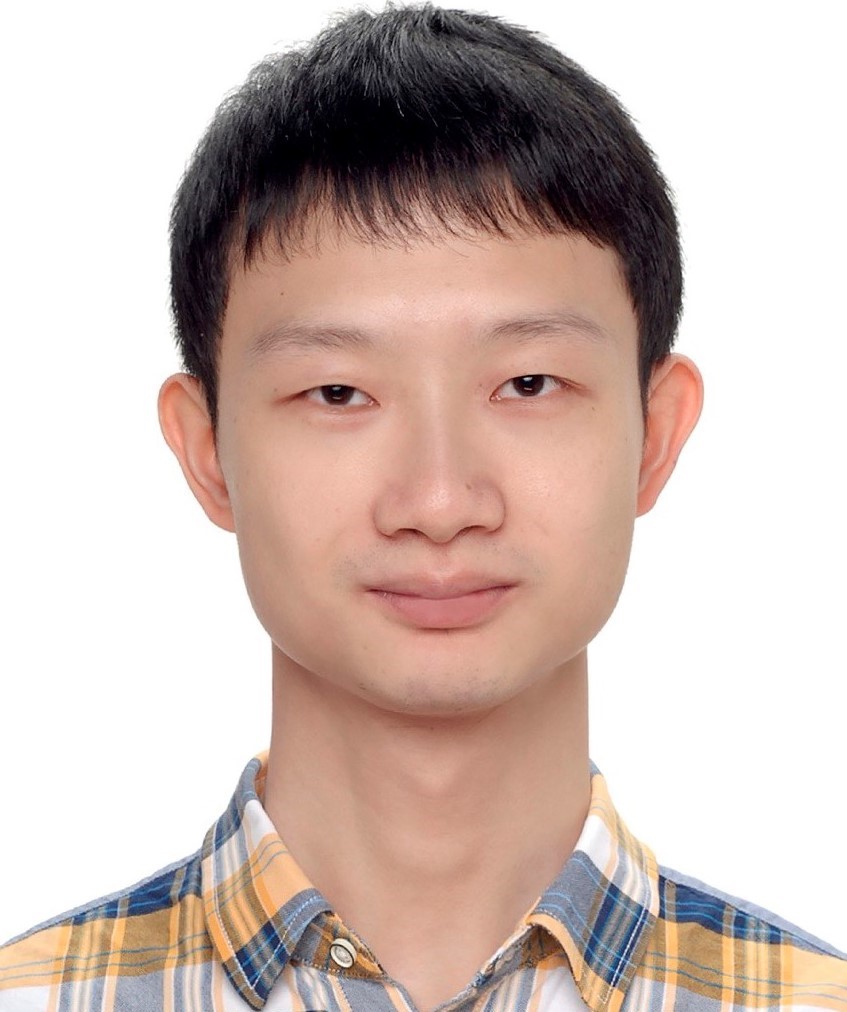}}]{Zhuosheng Zhang}
  		received his Bachelor's degree in internet of things from Wuhan University in 2016, his M.S. degree in computer science from Shanghai Jiao Tong University in 2020. He is working towards the Ph.D. degree in computer science with the Center for Brain-like Computing and Machine Intelligence of Shanghai Jiao Tong University. He was an internship research fellow at NICT from 2019-2020. His research interests include natural language processing, machine reading comprehension, dialogue systems, and machine translation. 
	\end{IEEEbiography}

\vspace{-12mm}
	\begin{IEEEbiography}[{\includegraphics[width=1in,height=1.25in,clip,keepaspectratio]{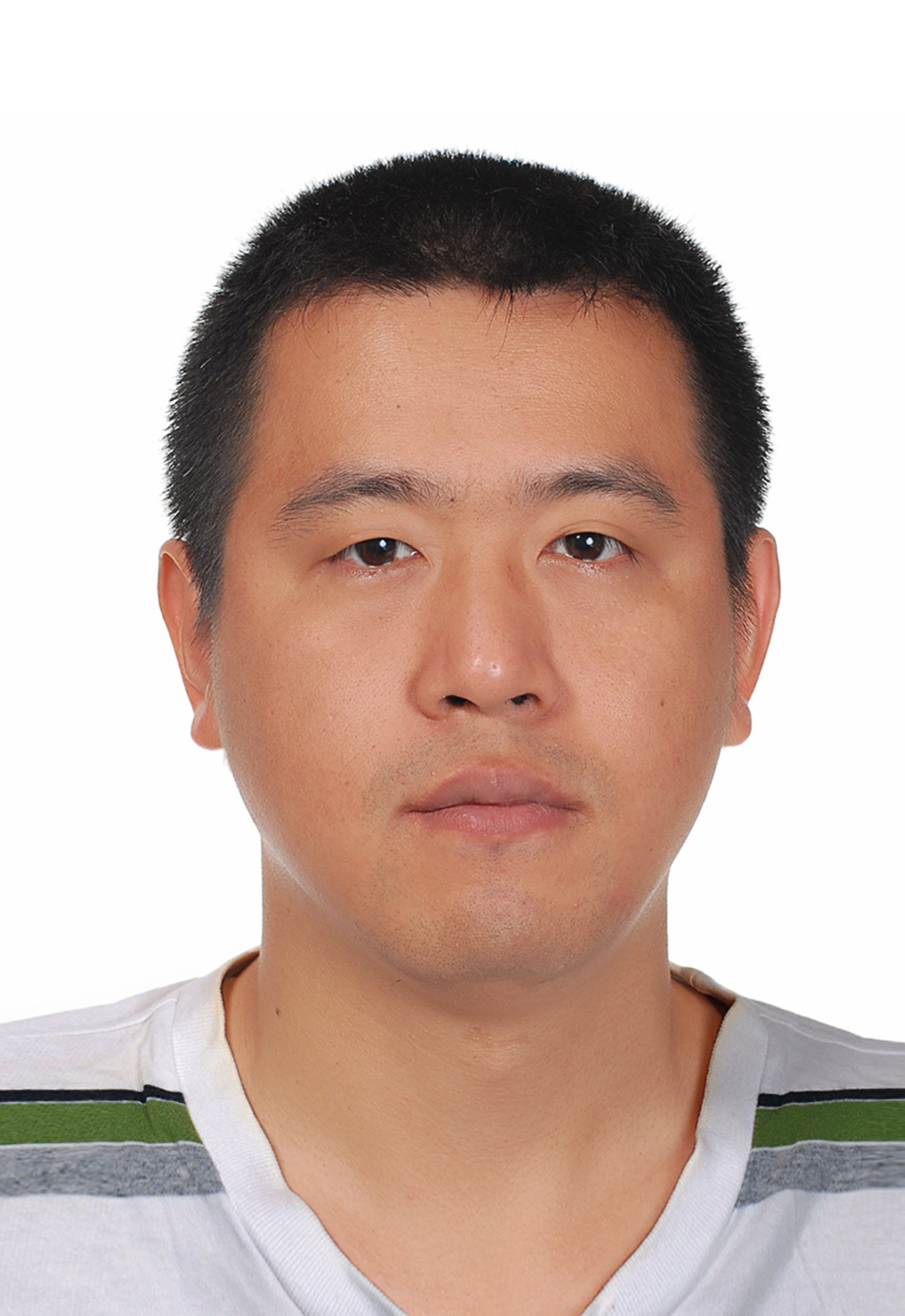}}]{Hai Zhao}
		received the BEng degree in sensor and instrument engineering, and the MPhil degree in control theory and engineering from Yanshan University in 1999 and 2000, respectively,
		and the PhD degree in computer science from Shanghai Jiao Tong University, China in 2005. 
		He is currently a full professor at department of computer science and engineering,  Shanghai Jiao Tong University after he joined the university in 2009. 
		He was a research fellow at the City University of Hong Kong from 2006 to 2009, a visiting scholar in Microsoft Research Asia in 2011, a visiting expert in NICT, Japan in 2012.
		He is an ACM professional member, and served as area co-chair in ACL 2017 on Tagging, Chunking, Syntax and Parsing, (senior) area chairs in ACL 2018, 2019 on Phonology, Morphology and Word Segmentation.
		His research interests include natural language processing and related machine learning, data mining and artificial intelligence.
	\end{IEEEbiography}

\end{document}